\documentclass[letterpaper, 10 pt, conference]{ieeeconf}  % Comment this line out if you need a4paper

\IEEEoverridecommandlockouts                              % This command is only needed if 
\usepackage[T1]{fontenc}
\usepackage{graphicx}
\usepackage{epsfig}
\usepackage{amsmath,amsfonts,amssymb,gensymb}
\usepackage{bm}
\usepackage{color}
\usepackage{float}
\usepackage{CJKutf8}
\usepackage{pifont}
\usepackage{cite}

\definecolor{forestgreen}{rgb}{0.13, 0.55, 0.13}

\definecolor{bluegray}{rgb}{0.4, 0.6, 0.8}

\title{\LARGE \bf
Design and Control of an Actively Morphing Quadrotor with Vertically Foldable Arms
}

\author{Tingyu Yeh$^{1}$*, Mengxin Xu$^{1}$* and Lijun Han$^{1}$% <-this % stops a space
\thanks{*These authors contributed equally to this work.}
\thanks{
This work was supported in part by the Natural Science Foundation of China under  62403311, in part by China Postdoctoral Science Foundation 2025T180468, GZC20231590 and 2024M761972.(\emph{Corresponding Author: Lijun Han})
}% <-this % stops a space
\thanks{$^{1}$Tingyu Yeh, Mengxin Xu and Lijun Han are with School of Automation and Intelligent Sensing, Shanghai Jiao Tong University and State Key Laboratory of Avionics Integration and Aviation System-of-Systems Synthesis, Shanghai 200240, China.(email: lanny\_yeh@sjtu.edu.cn, mengxin\_xu@sjtu.edu.cn, lijun\_han@sjtu.edu.cn)
}
}

\begin{document}

\maketitle
\thispagestyle{empty}
\pagestyle{empty}

%%%%%%%%%%%%%%%%%%%%%%%%%%%%%%%%%%%%%%%%%%%%%%%%%%%%%%%%%%%%%%%%%%%%%%%%%%%%%%%%
\begin{abstract}
% In this work, we propose a novel quadrotor design capable of folding its arms vertically to grasp objects and navigate through narrow spaces. The arms are actively folded using a central servomotor, gears, and racks, and they are structured as a parallelogram to connect the motor bases to the central frame. This design ensures that the orientation of the propellers remains unchanged during arm movement. 
% In its stretched state, the quadrotor resembles a conventional design, but when contracted, it takes on the appearance of a gripper, with grasping components emerging from the motor bases. To mitigate disturbances during the transformation and grasping processes, we employ an adaptive sliding mode controller alongside a disturbance observer. After fully folding, the quadrotor frame can be reduced to 67\% of its original size. The control performance and versatility of the morphing quadrotor are validated through real-world experiments.

In this work, we propose a novel quadrotor design capable of folding its arms vertically to grasp objects and navigate through narrow spaces. 
The transformation is controlled actively by a central servomotor, gears, and racks.
The arms connect the motor bases to the central frame, forming a parallelogram structure that ensures the propellers maintain a constant orientation during morphing. 
In its stretched state, the quadrotor resembles a conventional design, and when contracted, it functions as a gripper with grasping components emerging from the motor bases. To mitigate disturbances during transforming and grasping payloads, we employ an adaptive sliding mode controller with a disturbance observer. After fully folded, the quadrotor frame shrinks to 67\% of its original size. 
The control performance and versatility of the morphing quadrotor are validated through real-world experiments.

\end{abstract}

%%%%%%%%%%%%%%%%%%%%%%%%%%%%%%%%%%%%%%%%%%%%%%%%%%%%%%%%%%%%%%%%%%%%%%%%%%%%%%%%

\section{Introduction}

In recent years, research about quadrotors has moved beyond traditional fixed-shape structures, leading to the development of morphable and reconfigurable designs, with the aim of enhancing aerial capabilities and adaptability to diverse environments \cite{meng2022review,xing2024review}. 
These innovations have opened up new possibilities for aerial robots, enabling them to interact more effectively with the surroundings.

Various studies have explored shape transformation mechanisms that allow the vehicle to adapt to different environments.
Inspired by origami mechanism, a foldable quadrotor is proposed in \cite{yang2019origami} using a central servomotor to expand and contract the vehicle's body by changing the arm length.
Designs in \cite{riviere2018agile,desbiez2017xmorf} also employ a servomotor at the center, while the arms are rotated to reduce the wingspan.
In \cite{patnaik2020squeeze}, the quadrotor transforms by external forces when flying through narrow passages, and an improved learning-based controller is proposed for this completely passive morphing quadrotor in \cite{patnaik2022learning}.
% 這種主動或被動調整自身尺寸的能力，在zhao2021comparative 和cui2024motion 中得到驗證，能根據環境平衡靈活性和穩定性，或是在複雜環境中通過 變形能力獲得更優的飛行路徑。
Further validations have shown that such transformations can balance flexibility and stability according to the environment, or optimize flight paths in complex surroundings, as discussed in \cite{zhao2021comparative} and \cite{cui2024motion}.

% 更多工作在變形收縮尺寸的同時強調抓取物體或棲息的能力
More morphable designs emphasize object grasping or perching, further extending their capabilities in aerial manipulation. 
Compared with traditional aerial grasping missions, for example in \cite{zhang2018grasp,thomas2014avian}, the use of a morphable vehicle body reduces weight and complexity by eliminating the need for attached mechanical hands or grippers. 
In \cite{falanga2018foldable}, the quadrotor grasps objects through transformation using four independently controlled servomotors to rotate the arms around its body.
Similar frame designs can be found in \cite{kim2021morphing,cui2024motion,hu2021design}, where morphology is exploited for further applications such as optimized transportation.
In \cite{zhao2018deformable,wu2023ringrotor}, the quadrotors grasp objects by adjusting the central opening of their bodies through shape adaptation.
% Deformable arms are proposed for vehicle perching in \cite{ruiz2022sophie} and \cite{tao2023whopper}, utilizing soft materials and multi-link structures respectively. 
In \cite{jia2023modular, bucki2022hinges, bucki2019design}, elastic components are used in the frame, allowing the arms to fold automatically and rapidly when low thrusts are generated. These mechanical designs offer fast response and low weight by eliminating additional morphing actuators, but also introduce output constraints due to the reliance on propeller thrust for transformation. 
In contrast, a morphing quadrotor with grasping capabilities is proposed in \cite{Xu2024biomimetic}, which can actively fold its arms while maintaining stable flight. 

\begin{figure}[t]
    \centering
    \includegraphics[width=0.5\textwidth]{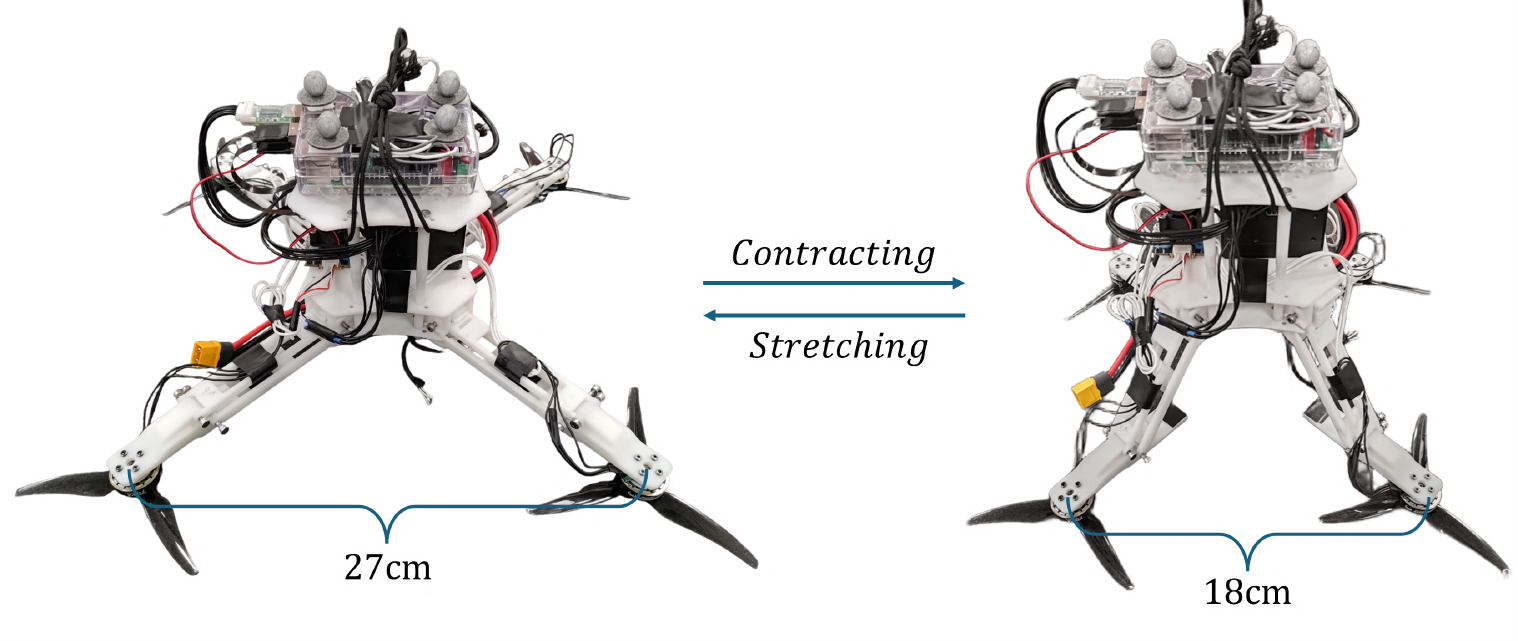}
    \caption{
            The prototype of our morphing quadrotor, with a stretched frame length of 27cm and a contracted length of 18cm. The rotors remain facing downward when the arms are folded. }
    \label{fig:real_body}
\end{figure}

In this paper, we present a novel quadrotor design that can fold its arms vertically to grasp objects and maneuver through confined spaces, as shown in Fig. \ref{fig:real_body}.
The proposed design uses 12 fewer links, reduces the overall weight by 10\%, and adjusts the contraction ratio from 80\% to 67\% compared to our previous work in \cite{Xu2024biomimetic}.
In addition, the quadrotor's gripper is integrated into the motor bases and directly connected to the ends of the arms via a parallelogram linkage structure. When the arms are unfolded, the gripper is concealed, making the quadrotor resemble a traditional one. When the central servomotor is activated to fold the arms vertically, the gripper becomes exposed. The parallelogram linkage ensures that the propellers remain oriented straight down throughout the morphing process. As a result, the quadrotor maintains stability in any configuration in a specified range, without sacrifice in thrust. However, proximity of the propellers when the arms are folded introduces aerodynamic disturbances, which affect the system's performance. To address this issue, we incorporate a disturbance observer into the controller for compensation. The control performance and versatility of the quadrotor are demonstrated through real-world experiments.

\section{System Overview}

\subsection{Quadrotor Design}

\begin{figure}[t]
    \centering
    \includegraphics[width=\linewidth]{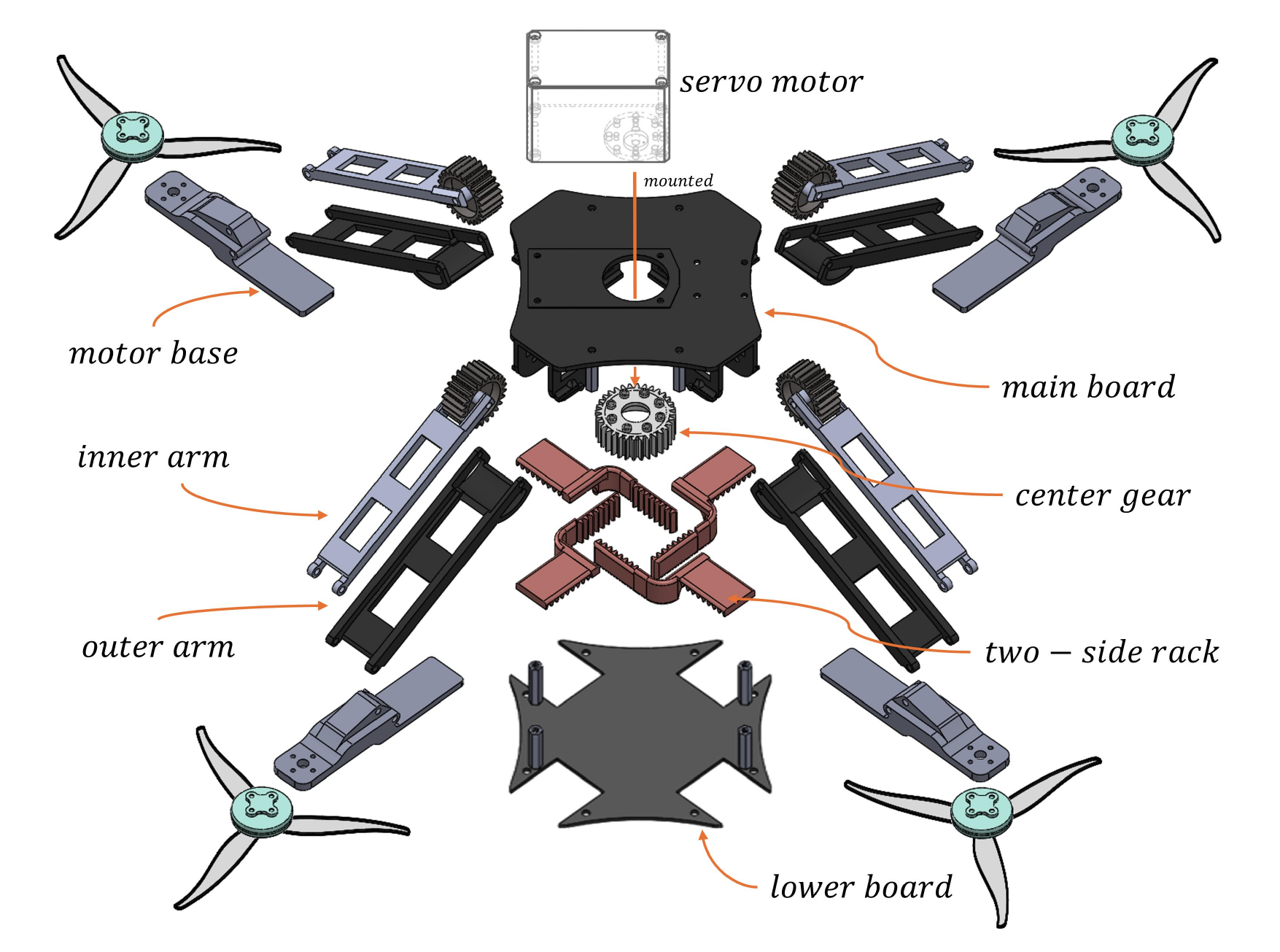}
    \caption{Overview of the quadrotor design from an exploded view.}
    \label{fig:assembly_demo}
\end{figure}
\begin{figure}[t]
    \centering
    \includegraphics[width=\linewidth]{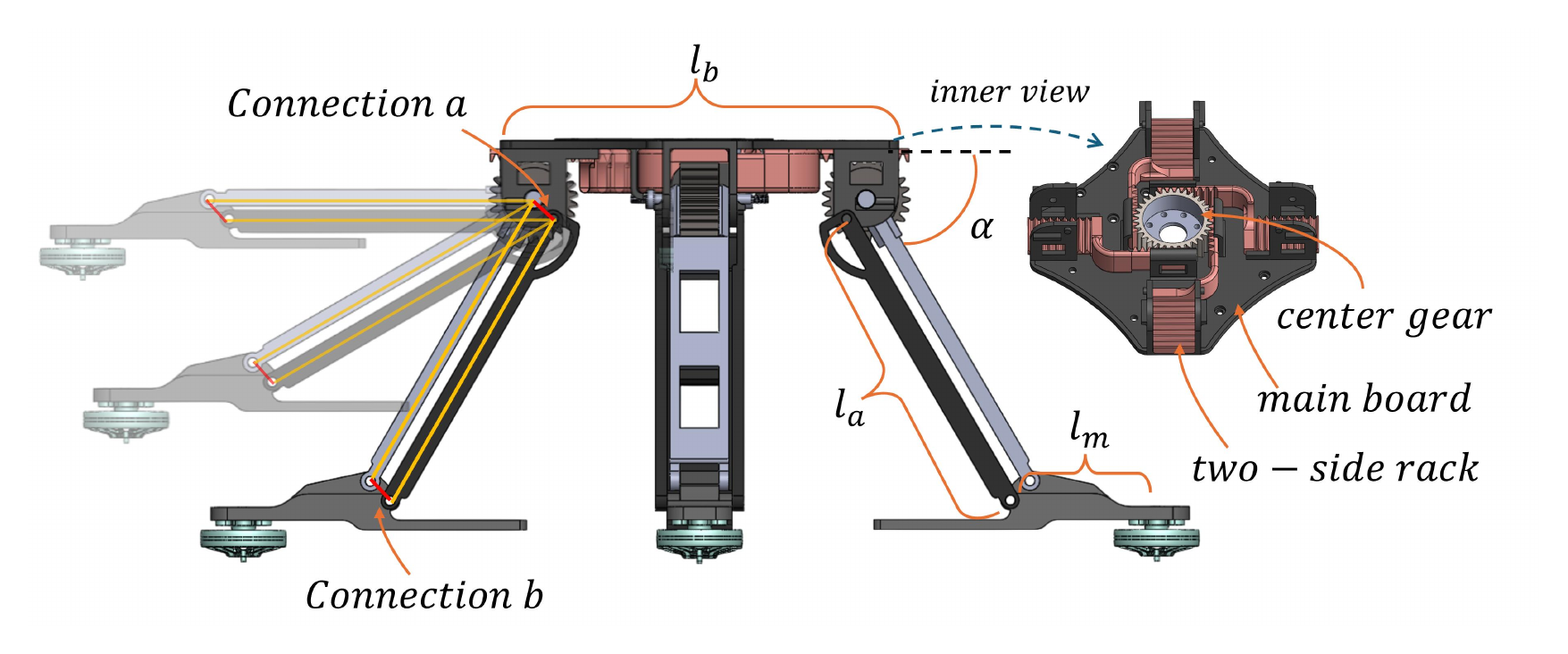}
    \caption{Front view of the morphing quadrotor. The connections \emph{a} and \emph{b}, indicated in red lines, maintain a fixed orientation during arm rotation, ensuring the rotors always face downward. The right side of the figure highlights the center gear and racks.}
    \label{fig:length_note}
\end{figure}

The quadrotor frame consists of a central body, four arms, and four motor bases, while the central body contains a servomotor, a gear, and four racks that actuate the morphing mechanism. All components of the frame are shown in Fig. 2.
Each arm comprises two equal-length links, with one end connected to the central body and the other attached to a motor base, indicated as connections \emph{a} and \emph{b} in Fig. 3 respectively.
The opposing sides of the connection are also of equal length, forming a parallelogram structure between the body, motor base, and two arm links.

During transformation, all four arms contract or extend simultaneously. By controlling the servomotor mounted on the central gear, the arms rotate vertically through the transmission of a specially designed two-sided rack, as illustrated on the right in Fig. \ref{fig:length_note}. 
Due to the geometric constraints of the parallelogram structure, the motor bases maintain a fixed orientation during arm rotation, ensuring that the rotors always face downward. 
This feature guarantees a consistent thrust direction and stable control performance, allowing a continuous 70-degree folding range for the arms. The motor bases are designed with overlapping sections with the arms when unfolded, and these sections function as a gripper during transformation, as they remain horizontal while the arms tilt.

\subsection{Modeling}

In this work, non-bold symbols represent scalars (e.g., m), bold lowercase letters represent vectors (e.g., $\boldsymbol{g}$), and bold uppercase letters are for matrices (e.g., $\boldsymbol{J}$). 
The skew-symmetric matrix for vector $\boldsymbol{a}$ is represented as $S(\boldsymbol{a})$.
We use $\{F_i\}$ to denote the inertial frame with its $z$ axis pointing upward, $\{F_b\}$ to denote the body frame defined at the center of gravity (CoG), and $\{F_c\}$ to represent the center frame located at the geometric center to calculate the CoG shift due to transformation.
% \begin{enumerate}
%     \item The inertial frame, denoted as $\{F_i:O_i - x_iy_iz_i\}$, with $z_i$ pointing upward, opposite to the gravity direction.
%     \item The body frame, denoted as $\{F_b:O_b - x_by_bz_b\}$ with the origin at the center of gravity(CoG).
%     \item The center frame $\{F_c:O_c - x_cy_cz_c\}$, located at the geometric center to calculate the CoG shifting due to transformation.
% \end{enumerate}
% In this section, we mainly describe the calculation of physical model that varies with the arms folding.
In this subsection, we mainly describe the calculations associated with the physical model, which changes dynamically when the arms rotate.

\subsubsection{Tip-to-Tip Length}

The front view of the quadrotor body is shown in Fig. \ref{fig:length_note}, 
where $l_b$, $l_a$ and $l_m$ stand for the
length of body, arms, and motor bases respectively. We denote the folding angle of arms as $\alpha$. Therefore, the tip-to-tip length $L$ of the quadrotor frame can be calculated as
\begin{equation}
    L = l_b+2l_a\cos{\alpha}+2l_m.
\end{equation}
%where $l_b$ represents the diagonal length of the main frame. 
The folding angle $\alpha$ is proportional to the rotation angle $\gamma$ of the central servomotor.

\subsubsection{Center of Gravity and Moment of Inertia}

As the arms fold, the CoG shifts and the moment of inertia changes according to the real-time configuration. 
The CoG offset $\boldsymbol{r}_{CoG}$ expressed in frame \{$F_c$\} yields to
\begin{equation}
    \boldsymbol{r}_{CoG} = 
    \frac{
    \sum_{i=1}^4
    (m_a\boldsymbol{r}_{ai} + 
    m_m\boldsymbol{r}_{mi} + 
    m_r\boldsymbol{r}_{ri})
    }
    {m_b+4(m_a+m_m+m_r)},
\end{equation}
where $\boldsymbol{r}_{ai}$, $\boldsymbol{r}_{mi}$, $\boldsymbol{r}_{ri}$
indicate the CoG position of $i$-th arm, motor base, rotor, and $m_{a}$, $m_{m}$, $m_{r}$ denote the mass of them. $m_{b}$ is the mass of the central body.

For the calculation of moment of inertia $\boldsymbol{J}$, we first approximate the modules as regular geometric shapes. 
The central body, along with the attached electronic components such as the servomotor and batteries, is modeled as a box with height $h_b$. 
The rotors are modeled as cylinders with height $h_r$ and radius $r_r$.
The arms and motor bases are represented as rectangular cuboids.
Therefore, we have these modules' moment of inertia calculated as follows:
\begin{align}
    \boldsymbol{J}_b &= \frac{m_b}{12} * diag(l_b^2+h_b^2, l_b^2+h_b^2, l_b^2+l_b^2), \\ \notag
    \boldsymbol{J}_r &= \frac{m_r}{12} * diag(3r_r^2+h_r^2,3r_r^2+h_r^2,6h_r^2).
\end{align}
The calculations for $\boldsymbol{J}_m$ and $\boldsymbol{J}_a$ are similar to that of $\boldsymbol{J}_b$. Meanwhile, due to the rotation of arms, $\boldsymbol{J}_a$ varies with the folding angle $\alpha$ as
\begin{equation}
    \boldsymbol{J}_{ai} = \boldsymbol{R}_{ai}(\alpha) \boldsymbol{J}_a \boldsymbol{R}_{ai}(\alpha)^\intercal,
\end{equation}
where $\boldsymbol{R}_{ai}(\alpha)$ is the rotation matrix.
The moment of inertia $\boldsymbol{J}$ of the system is derived as follows:
\begin{equation}
    \begin{aligned}
            &\boldsymbol{J} = \boldsymbol{J}_b - m_bS(-\boldsymbol{r}_{CoG})^2  
                    + \sum_{i=1}^4 (\boldsymbol{J}_{ai} - m_aS(\boldsymbol{r}_{ai} - \boldsymbol{r}_{CoG})^2 \\ 
                    &+ \boldsymbol{J}_{mi} - m_mS(\boldsymbol{r}_{mi} - \boldsymbol{r}_{CoG})^2   
                    + \boldsymbol{J}_{ri} - m_rS(\boldsymbol{r}_{ri} - \boldsymbol{r}_{CoG})^2).
    \end{aligned}
\end{equation}
Since the whole vehicle is designed and assembled to be symmetric, $\boldsymbol{J}$ is a diagonal matrix and can be expressed as $\boldsymbol{J}=diag(J_{xx},\,\,J_{yy},\,\,J_{zz})$.

\subsection{Dynamics}

In this subsection, we present translational and rotational dynamics of the quadrotor. With $\boldsymbol{p} = [ x\,\,\, y\,\,\, z]^\intercal$ denoting the position of the vehicle in inertial frame \{$F_i$\}, 
and $\boldsymbol{\zeta} = [\phi\,\,\, \theta\,\,\, \psi]^\intercal$ denoting the attitude in Euler angle form,
the dynamics are expressed as
\begin{align}
    & m \Ddot{\boldsymbol{p}} + m \boldsymbol{g} = \boldsymbol{f} + \boldsymbol{f}_d, \label{eq:trans_dynamics}
    \\ 
    \boldsymbol{B}(\boldsymbol{\zeta})\Ddot{\boldsymbol{\zeta}} &+ \boldsymbol{C}(\Dot{\boldsymbol{\zeta}},\boldsymbol{\zeta})\Dot{\boldsymbol{\zeta}} = \boldsymbol{\tau} + \boldsymbol{\tau}_d,
\end{align}
in which $m$ is the system mass and $\boldsymbol{g}$ is the gravity acceleration. 
$\boldsymbol{B}(\boldsymbol{\zeta})$ is the moment of inertia described in frame \{$F_i$\}, 
and $\boldsymbol{C}(\Dot{\boldsymbol{\zeta}},\boldsymbol{\zeta})$ represents the Coriolis matrix. 
In addition, $\boldsymbol{f}$ and $\boldsymbol{\tau}$ denote the control force and torque, while $\boldsymbol{f}_d$ and $\boldsymbol{\tau}_d$ denote the disturbances, which are assumed to be bounded with maximum values of $\boldsymbol{f}_m$ and $\boldsymbol{\tau}_m$ respectively. 

\begin{figure}
    \centering
    \includegraphics[width=0.95\linewidth]{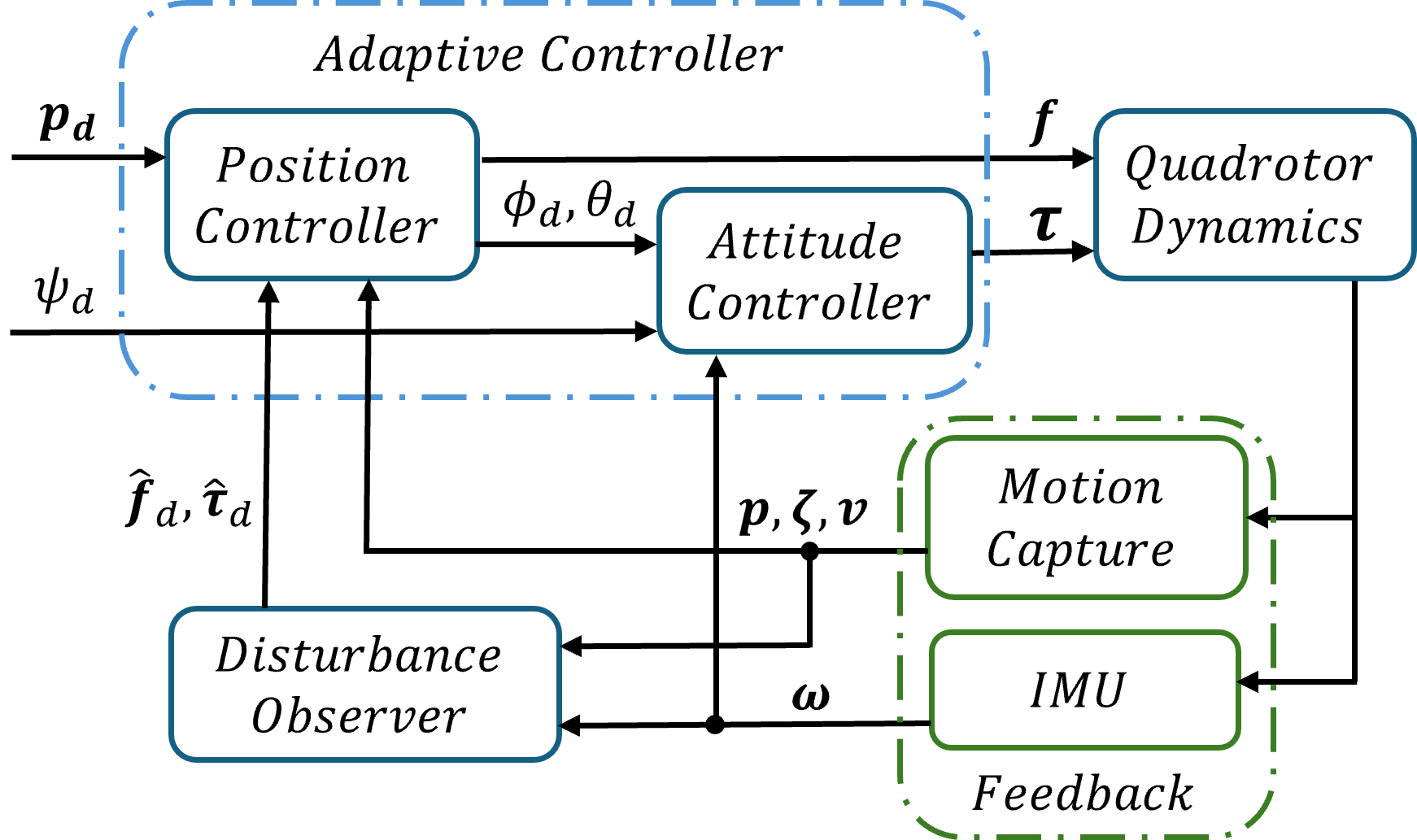}
    \caption{
        Structure of the proposed cascaded controller. $\boldsymbol{p}_d, \psi_d$: desired position and yaw angle. $\phi_d,\theta_d$: desired roll and pitch angle. $\boldsymbol{f},\boldsymbol{\tau}$: calculated force and torque. $\boldsymbol{p},\boldsymbol{\zeta},\boldsymbol{v},\boldsymbol{\omega}$: measured position, attitude, velocity and angular velocity. $\widehat{\boldsymbol{f}}_d,\widehat{\boldsymbol{\tau}}_d$: estimated disturbance force and torque.
    }
    \label{fig:control_structure}
\end{figure}

% The Euler angle rates $\Dot{\boldsymbol{\zeta}}$ can be converted from the angular velocity $\boldsymbol{\omega} = [ r\,\,\, p\,\,\, y ]$ in body frame, with the transformation
% \begin{equation}
%     \Dot{\boldsymbol{\zeta}} = \boldsymbol{Q}^{-1}\boldsymbol{\omega},
% \end{equation}
% where $\boldsymbol{Q}$ is a matrix related to current attitude, given by
% \begin{equation}
%     \boldsymbol{Q}^{-1} = 
%     \begin{pmatrix}
%         1 & \sin(\phi)\tan(\theta) & \cos(\phi)\tan(\theta)\\
%         0 & \cos(\phi)             & -\sin(\phi)           \\
%         0 & \frac{\sin(\phi)}{\cos(\theta)}  & \frac{\cos(\phi)}{\cos(\theta)}.
%     \end{pmatrix}
% \end{equation}
% Finally, $f$ and $\tau$ depict the desired control force and torque, while $f_d$ and $\tau_d$ denote the disturbance ones, with the bounded maximum of $f_m$ and $\tau_m$ respectively. 

\section{Control Strategy}

Benefiting from the parallelogram structure design of the arms, the fixed orientation of the propellers allows the system to use the same controller throughout the entire transformation process. However, variations in moment of inertia and total mass during morphing and object grasping should still be taken into account. 
Furthermore, when the arms are folded, the reduced distance between the propellers amplifies the impact of aerodynamic disturbances on the system. To mitigate this, we build on the adaptive sliding mode controller from our previous work \cite{Xu2024biomimetic} by incorporating an observer to further compensate for the disturbance force and torque. The overview of the proposed control strategy is shown in Fig. \ref{fig:control_structure}.

\subsection{Disturbance Observer}

In this subsection, we introduce the design of the disturbance observer. The observer estimates the disturbance force and torque, which are then used in the controller for feed-forward compensation.
The design follows the method proposed in \cite{Ruggiero2014Impedance}, and a similar application can be found in \cite{Bodie2021Active}.

For convenience, the disturbance torque observer is designed in the body frame $\{F_b\}$ with the rotational dynamics described as 
\begin{align}
    \boldsymbol{J}\dot{\boldsymbol{\omega}}+
    S(\boldsymbol{\omega}) (\boldsymbol{J}\boldsymbol{\omega}) 
    = \boldsymbol{\tau}^b
    +\boldsymbol{\tau}_{d}^b. \label{eq:body dynamics}
\end{align}
$\boldsymbol{\tau}^b$ and $\boldsymbol{\tau}_{d}^b$ are the control and disturbance torque respectively. 

The real-time estimation of disturbances is given by:
\begin{align}
    \widehat{\boldsymbol{f}}_d &=
    \boldsymbol{K}_f 
    (m\dot{\boldsymbol{p}} - 
    \int(
    \boldsymbol{f}_c -m\boldsymbol{g}+\widehat{\boldsymbol{f}}_{d})dt),
    \\
    \widehat{\boldsymbol{\tau}}_{d}^b &= 
    \boldsymbol{K}_t 
    (\boldsymbol{J}\boldsymbol{\omega}- 
    \int (\boldsymbol{\tau}_c-S(\boldsymbol{\omega}) (\boldsymbol{J}\boldsymbol{\omega})+
    \widehat{\boldsymbol{\tau}}_{d}^b)dt),
\end{align}
where $\boldsymbol{K}_f$ and $\boldsymbol{K}_t$ are diagonal matrices for positive gains. 
$\boldsymbol{f}_c$ and $\boldsymbol{\tau}_c$ denote the command force and torque, which are assumed to be able to achieve the actual applied $\boldsymbol{f}$ and $\boldsymbol{\tau}$.
Combining the derivative of the above equations and the dynamics together, we obtain the closed-loop equations of the observer as
\begin{align}
    \dot{\widehat{\boldsymbol{f}}}_d &=
    \boldsymbol{K}_f
    (\boldsymbol{f}_d - \widehat{\boldsymbol{f}}_d), \\
    \dot{\widehat{\boldsymbol{\tau}}}^b_{d} &= 
    \boldsymbol{K}_t 
    (\boldsymbol{\tau}_{d}^b -
    \widehat{\boldsymbol{\tau}}_{d}^b ),
\end{align}
% which can be viewed as a first-order low-pass filter system.
where the estimate of the external disturbance is equivalent to the actual value passed through a first-order low-pass filter.

\subsection{Cascade Adaptive Controller}

The proposed controller features a classical cascade structure, consisting of a position loop and an attitude loop. 
The controller first takes the desired position $\boldsymbol{p}_d$ to compute the target force $\boldsymbol{f}$, then combines the desired heading direction $\psi_d$ with $\boldsymbol{f}$ to obtain the target Euler angle $\boldsymbol{\zeta}_d$. 
Afterward, the desired torque $\boldsymbol{\tau}$ is derived from the attitude loop.
Finally, the estimated disturbance force and torque obtained from the observer are incorporated into the controller for compensation.

The controller design starts by defining tracking errors as follows:
\begin{align}
    \boldsymbol{e}_1=\boldsymbol{p}-\boldsymbol{p}_d,\,\,
    \boldsymbol{e}_2=\boldsymbol{\zeta}-\boldsymbol{\zeta}_d.
\end{align}
And the sliding surfaces are chosen as
\begin{align}
\boldsymbol{s}_{1} &
    =\dot{\boldsymbol{e}}_{1}+\boldsymbol{\Lambda}_{1}\boldsymbol{e}_{1}
    =\dot{\boldsymbol{p}}-\dot{\boldsymbol{p}}_{r},
    \\
    \boldsymbol{s}_{2} &
    =\dot{\boldsymbol{e}}_{2}+\boldsymbol{\Lambda}_{2}\boldsymbol{e}_{2}
    =\dot{\boldsymbol{\zeta}}-\dot{\boldsymbol{\zeta}}_{r},
\end{align}
with
\begin{align}
    \dot{\boldsymbol{p}}_r=
    \dot{\boldsymbol{p}}_d - \boldsymbol{\Lambda}_1\boldsymbol{e}_1,\,\,
    \dot{\boldsymbol{\zeta}}_r=
    \dot{\boldsymbol{\zeta}}_d - \boldsymbol{\Lambda}_2\boldsymbol{e}_2,
\end{align}
where $\boldsymbol{\Lambda}_1$, $\boldsymbol{\Lambda}_2$ are positive-definite diagonal matrices.

Since the moment of inertia $\boldsymbol{J}$ is a diagonal matrix, we extract a vector $\boldsymbol{b} =[J_{xx}\,\,\, J_{yy}\,\,\, J_{zz}]^\intercal$ from $\boldsymbol{J}$, then rewrite the rotational dynamics as
\begin{align}
    \boldsymbol{B}(\boldsymbol{\zeta})
    \ddot{\boldsymbol{\zeta}}_r
    +\boldsymbol{C}(\dot{\boldsymbol{\zeta}},\boldsymbol{\zeta})
    \dot{\boldsymbol{\zeta}}_r
    =\boldsymbol{Y}(\ddot{\boldsymbol{\zeta}}_r,\dot{\boldsymbol{\zeta}}_r,
    \dot{\boldsymbol{\zeta}},\boldsymbol{\zeta})
    \boldsymbol{b}.
\end{align}

In order to mitigate the oscillation issue of the sliding mode controller, we introduce saturation functions into sliding surfaces by creating new terms as follows:
\begin{align}
    \boldsymbol{\Delta}_1 = \boldsymbol{s}_1-\sigma_1 sat(\boldsymbol{s}_1/\sigma_1), \\
    \boldsymbol{\Delta}_2 = \boldsymbol{s}_2-\sigma_2 sat(\boldsymbol{s}_2/\sigma_2),
\end{align}
where $\sigma_1$ and $\sigma_2$ are positive parameters.

Based on the above definitions, the proposed control law is given by:
\begin{align}
    \boldsymbol{f} &=
    \widehat{m}(\boldsymbol{g}+\ddot{\boldsymbol{p}}_{r})-
    \boldsymbol{K}_{p1}\boldsymbol{\Delta}_{1}-
        \boldsymbol{K}_{p2}sat(\boldsymbol{s}_{1}/\sigma_{1}) - \widehat{\boldsymbol{f}}_d,\\
    \boldsymbol{\tau} &= 
    \boldsymbol{Y}\widehat{\boldsymbol{b}}-
    \boldsymbol{K}_{\zeta1}\boldsymbol{\Delta}_{2}
    -(\boldsymbol{K}_{\zeta2}-\sigma_{2}\boldsymbol{C})
    sat(\boldsymbol{s}_{2}/\sigma_{2}) - \widehat{\boldsymbol{\tau}}_d,
\end{align}
where all the $\boldsymbol{K}_{*}$ are diagonal matrices with positive values, while $\widehat{m}$ and $\widehat{\boldsymbol{b}}$ are the estimates of the mass $m$ and inertia vector $\boldsymbol{b}$. 
The estimated disturbance torque expressed in inertial frame can be converted from $\widehat{\boldsymbol{\tau}}_d^b$.

For the estimated value of $\widehat{m}$ and $\widehat{\boldsymbol{b}}$, the estimation errors are defined as:
\begin{align}
    \Tilde{m} = m-\widehat{m},\,\,
    \Tilde{\boldsymbol{b}} = 
    \boldsymbol{b} - \widehat{\boldsymbol{b}}.
\end{align}
The adaptive law is designed with positive gains 
$\Gamma_1$ and $\boldsymbol{\Gamma}_2$, given by:
\begin{align}
    \dot{\widehat{m}} &= 
    - \Gamma_1^{-1}
    (\boldsymbol{g}+\ddot{\boldsymbol{p}}_r)^\intercal
    \boldsymbol{\Delta}_1,
    \\
    \dot{\widehat{\boldsymbol{b}}} &=
    - \boldsymbol{\Gamma}_2^{-1}
    \boldsymbol{Y}^\intercal
    \boldsymbol{\Delta}_2.
\end{align}

With the above control law and adaptive law, we can prove the system's stability by choosing a Lyapunov candidate function as:
\begin{align}
    V = 
    \frac{1}{2}\boldsymbol{\Delta}_1^\intercal m \boldsymbol{\Delta}_1
    + \frac{1}{2}\Gamma_1\Tilde{m}^2
    + \frac{1}{2}\boldsymbol{\Delta}_2^\intercal \boldsymbol{B} \boldsymbol{\Delta}_2
    + \frac{1}{2}\Tilde{\boldsymbol{b}}^\intercal\boldsymbol{\Gamma}_2\Tilde{\boldsymbol{b}},
\end{align}
with its time derivative as:
\begin{equation}
    \begin{aligned}
    \dot{V} = -&\boldsymbol{\Delta}_1^\intercal \boldsymbol{K}_{p1} \boldsymbol{\Delta}_1
    - \boldsymbol{\Delta}_1^\intercal \boldsymbol{K}_{p2}sat(\boldsymbol{s}_1/\sigma_1)
    + \boldsymbol{\Delta}_1^\intercal (\boldsymbol{f}_d - \widehat{\boldsymbol{f}}_d)
    \\
    -&\boldsymbol{\Delta}_2^\intercal \boldsymbol{K}_{\zeta1} \boldsymbol{\Delta}_2
    - \boldsymbol{\Delta}_2^\intercal \boldsymbol{K}_{\zeta2}sat(\boldsymbol{s}_2/\sigma_2)
    + \boldsymbol{\Delta}_2^\intercal (\boldsymbol{\tau}_d - \widehat{\boldsymbol{\tau}}_d).
\end{aligned}
\end{equation}
Since $\bm{\Delta}_i^\intercal sat(\bm{s}_i/{\sigma}_i)=|\bm{\Delta}_i|$ and $\boldsymbol{f}_d, \boldsymbol{\tau}_d$ are assumed to be bounded, with $\boldsymbol{K}_{p2}$ and $\boldsymbol{K}_{\zeta2}$ chosen properly, we have $\dot{V} \leq 0$. More detailed proof of the stability can be found in \cite{Xu2024biomimetic}. 
It is worth noting that we actively reduce the disturbance term $\boldsymbol{f}_d$ and $\boldsymbol{\tau}_d$ with the observer, which enhances the tolerance to external disturbances.

% Furthermore, the total thrust $T$ is obtained by $ T = ||\boldsymbol{f}||$, and the desired Euler angle $\boldsymbol{\zeta}_d$ is derived from $\boldsymbol{f}$ and the command heading direction $\psi_d$. 
% We introduce an intermediate axis $\boldsymbol{y}_{tmp}$ with:
% \begin{align}
%     \boldsymbol{y}_{tmp} = [-\sin(\psi_d) \,\,\,\cos(\psi_d)\,\,\,0]^\intercal.
% \end{align}
% Given $\boldsymbol{z}_b = \boldsymbol{f}/T$, the desired $\boldsymbol{y}_b$ and $\boldsymbol{x}_b$ can be obtained by:
% \begin{align}
%     \boldsymbol{y}_b = \boldsymbol{z}_b \times \boldsymbol{x}_b,\,\,
%     \boldsymbol{x}_b = \frac{\boldsymbol{y}_{tmp} \times \boldsymbol{z}_b }{||\boldsymbol{y}_{tmp} \times \boldsymbol{z}_b||}.
% \end{align}
% With $[ \boldsymbol{x}_b \,\,\, \boldsymbol{y}_b \,\,\, \boldsymbol{z}_b]$ forms the target body attitude $\boldsymbol{R}_d$, the desired Euler angle $\boldsymbol{\zeta}_d$ can be extracted straightforwardly. More details about this method can be found in \cite{Lee2010Geometric}. 
\section{Experiments}

% In this section, we present the experimental results to validate the abilities of our morphing quadrotor and the performance of the controller design. 
% We first describe the hardware and physical properties of the vehicle. Then, we perform comparison experiments to verify the improvement by introducing the disturbance observer into the controller.
% Following that, several experiments are conducted including trajectory tracking, object grasping, and narrow-space maneuvering to show the versatility of our morphing quadrotor.

This section presents experimental results validating the capabilities of the morphing quadrotor and the performance of the controller. We first describe the hardware and physical properties, then perform comparative experiments to verify the improvement by introducing the disturbance observer into the controller.
Following that, the quadrotor’s versatility is demonstrated through trajectory tracking, object grasping, and narrow-space maneuvering experiments.

\subsection{Experiment Setup}
The experiments are performed with the prototype shown in Fig. \ref{fig:real_body}, which has a total mass of $0.805$ kg and dimensions of $l_b = 12.2$ cm, $l_a = 9$ cm, and $l_m = 4$ cm, as indicated in Fig. \ref{fig:length_note}. The frame length, defined as the distance between two adjacent motor bases, varies from a maximum of 27 cm to a minimum of 18 cm. A Raspberry Pi 4B is used as the companion computer to control the quadrotor and servo, while an Arduino Nano transmits rotor speed commands to the electronic speed controllers. Position and attitude data are provided by a Vicon system and an onboard IMU. Specific models of the electrical components can be found in \cite{Xu2024biomimetic}.

% The parameters of the proposed controller are listed in Table \ref{table:parameters}. \update{Why is the table number in this style...o\_o}

% \begin{table}[ht]
% \centering

% \label{table:parameters}
% \caption{Controller Parameters}

% \begin{tabular}{llll}
% \hline
% % \multicolumn{4}{c}{Parameters}                                                                                             \\ \hline
% \multicolumn{1}{l|}{$K_{p1}$}    & \multicolumn{1}{l|}{$diag[0.4,\,\,0.4,\,\,3]$} & \multicolumn{1}{l|}{$K_{\Phi 1}$} & $diag[0.07,\,\,0.07,\,\,0.1]$ \\ \hline
% \multicolumn{1}{l|}{$\Lambda_1$} & \multicolumn{1}{l|}{$diag[7.5,\,\,7.5,\,\,1.7]$} & \multicolumn{1}{l|}{$\Lambda_2$}  & $diag[7.1,\,\,7.1,\,\,1]$ \\ \hline
% \multicolumn{1}{l|}{$K_{p2}$}    & \multicolumn{1}{l|}{$0.01*diag[0.5,\,\,0.5,\,\,1]$} & \multicolumn{1}{l|}{$K_{\Phi 2}$} & $0.01* diag[0.1,\,\,0.1,\,\,1]$  \\ \hline
% \multicolumn{1}{l|}{$\Gamma_1$}  & \multicolumn{1}{l|}{0.015}                & \multicolumn{1}{l|}{$\Gamma_2$}   &  $diag[0,\,\,0,\,\,0]$     \\ \hline
% \multicolumn{1}{l|}{$\sigma_1$}  & \multicolumn{1}{l|}{0.03}                  & \multicolumn{1}{l|}{$\sigma_2$}   &   0.05     \\ \hline
% \multicolumn{1}{l|}{$K_f$}       & \multicolumn{1}{l|}{$diag[0.5,\,\,0.5,\,\,0]$} & \multicolumn{1}{l|}{$K_t$}    &    $diag[0.5,\,\,0.5,\,\,5]$  \\ \hline
% \end{tabular}%

% \end{table}

\begin{figure}[t]
    \centering
    \includegraphics[width=\linewidth]{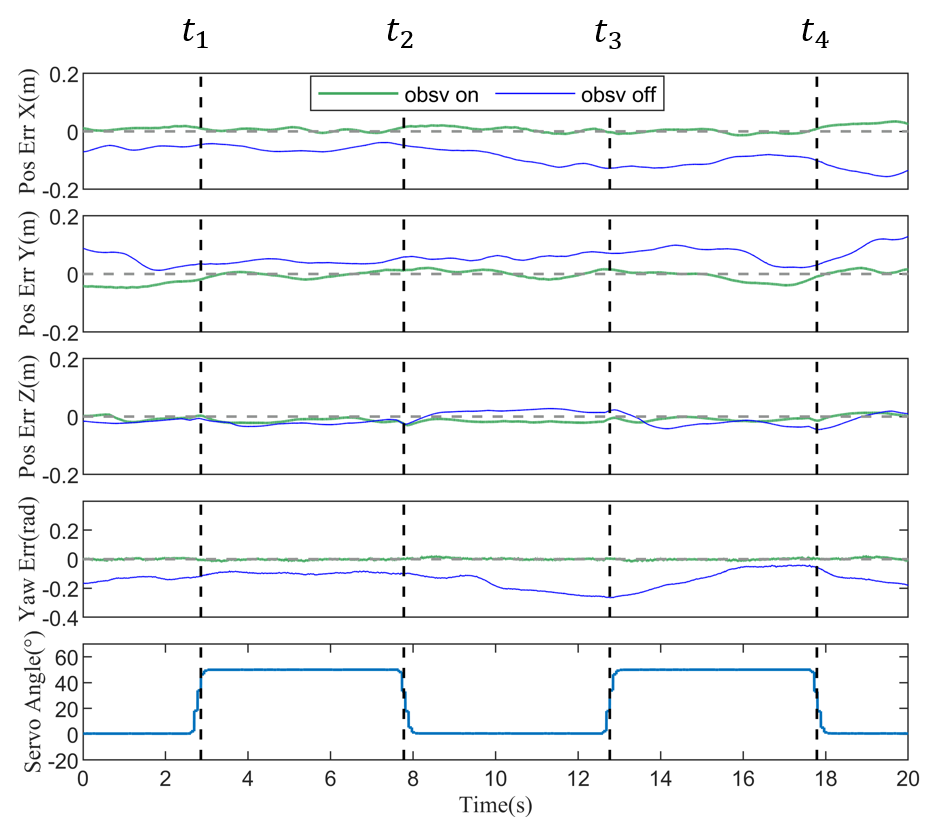}
    \caption{
    Results of the comparative experiments with the disturbance observer enabled and disabled, respectively. 
    The quadrotor contracted its frame at $t_1$ and $t_3$, and fully expanded at $t_2$ and $t_4$. 
    Tracking errors occurred when the observer was disabled. 
    }
    \label{fig:data_compare}
\end{figure}
\begin{figure}[!tbp]
    \centering
    \includegraphics[width=\linewidth]{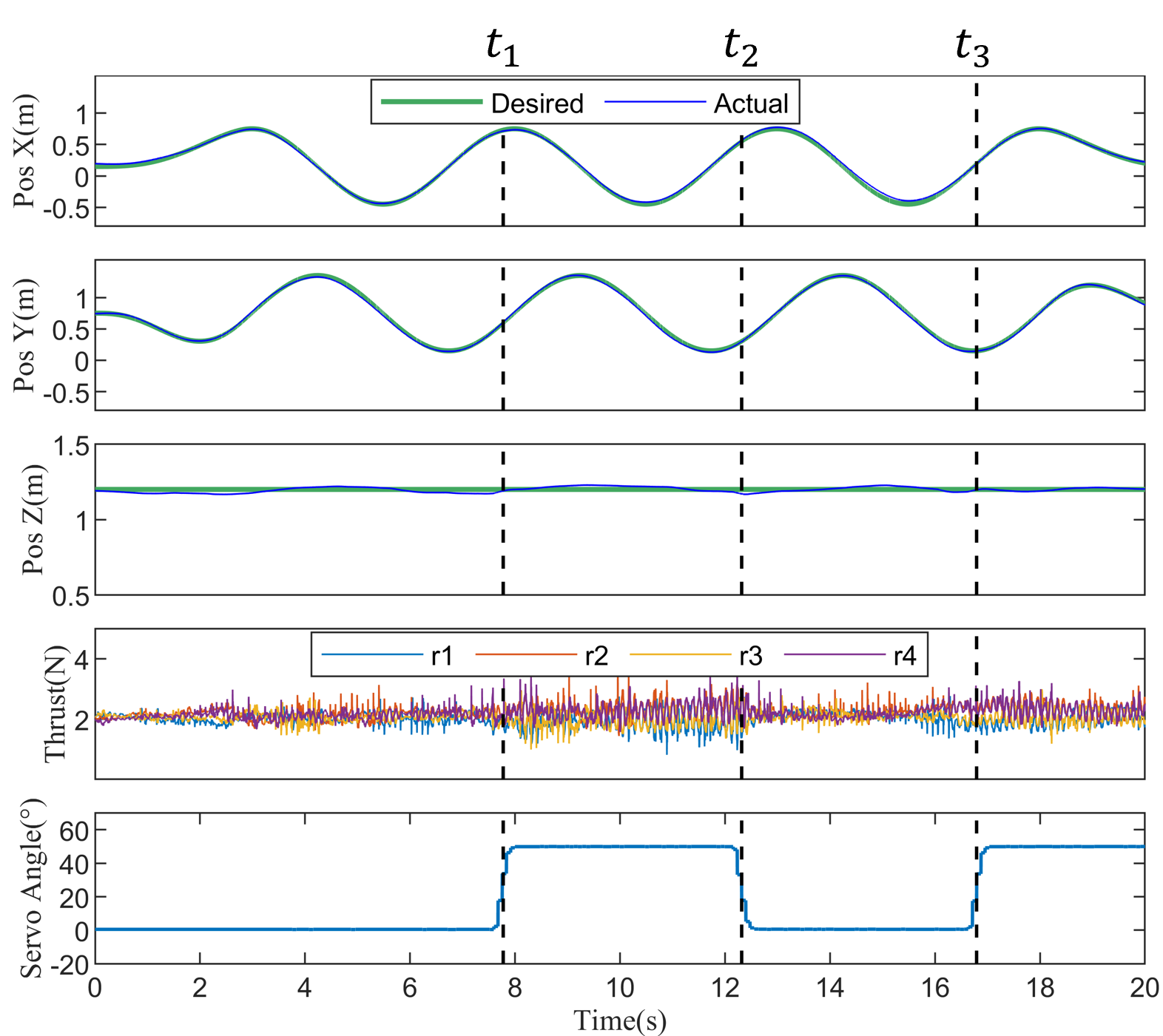}
    \caption{
        Results of the circular trajectory tracking. The quadrotor contracted at $t_1$, $t_3$ and expanded at $t_2$. 
        The results show no significant offset even during the morphing process. The tracking error remained within 0.05 m.
        }
    \label{fig:data_circle_morphing}
\end{figure}

\subsection{Comparison of Disturbance Observer On vs. Off}

To validate the enhancement provided by the introduced disturbance observer, we conduct comparative experiments using the controller with the observer enabled and disabled, 
by having the quadrotor hover while periodically changing its configuration.
Fig. \ref{fig:data_compare} presents the position tracking errors of the two hovering experiments.
The servo angle changed from $0 \degree$ to $50 \degree$ for minimal contraction, then returned to $0 \degree$ for full expansion. 
Contraction occurred at $t_1 = 2.85\,\text{s}$ and $t_3 = 12.85\,\text{s}$, while expansion was applied at $t_2 = 7.85\,\text{s}$ and $t_4 = 17.85\,\text{s}$.  
The results show that 
drift and oscillation occurred when the disturbance observer was disabled. In contrast, the controller can compensate for the disturbances when enabling the observer, improving tracking accuracy and maintaining the quadrotor's stability.

\subsection{Actively Morphing in Trajectory Tracking}

We further demonstrate the controller's performance by morphing while tracking a circular trajectory with a radius of $0.6\,\text{m}$, centered at $(x,\,y) = (0,\,0.6)\,\text{m}$, at a fixed flight altitude of $1.2\,\text{m}$. 
The quadrotor completes one lap in 5 seconds. 
As depicted in Fig. \ref{fig:data_circle_morphing},
it started tracking in the stretched mode and changed the configuration at $t_1 = 7.7\,\text{s}$, $t_2 = 12.7\,\text{s}$ and $t_3 = 16.7\,\text{s}$.
The results show no significant offset
even during the morphing process. 
Despite repeated configuration changes, the proposed controller effectively maintains the tracking error within $0.05\,\text{m}$, validating its reliability and stable performance.

\subsection{Grasping Object and Transporting}

In this experiment, our morphing quadrotor demonstrates its ability to grasp and transport objects.
% As previously described, the gripper is integrated into the motor bases, enabling grasping and releasing through arm folding and unfolding. 
% Therefore, the operations of grasping and releasing can be performed by folding and unfolding the arms, which drive the gripper to close and open.
The complete transportation procedure is shown in Fig. \ref{fig:video-grasp}. 
% The target box has the dimensions of $12 \times 15 \times 19\,\text{cm}$ and weighs $77\,\text{g}$. However, its mass is unknown to the controller during the grasping experiments.
The target box measures $12 \times 15 \times 19\,\text{cm}$ and weighs $77\,\text{g}$, but its mass is unknown to the controller during the experiment. 
The quadrotor initially hovered in stretched mode above the target box before folding its arms to contract the grippers for grasping. Subsequently, it maneuvered in folded configuration while carrying the payload. Upon reaching the planned position, the quadrotor stretched its body to release the gripper and drop the box.

\begin{figure}[!htbp]
    \centering
    \includegraphics[width=\linewidth]{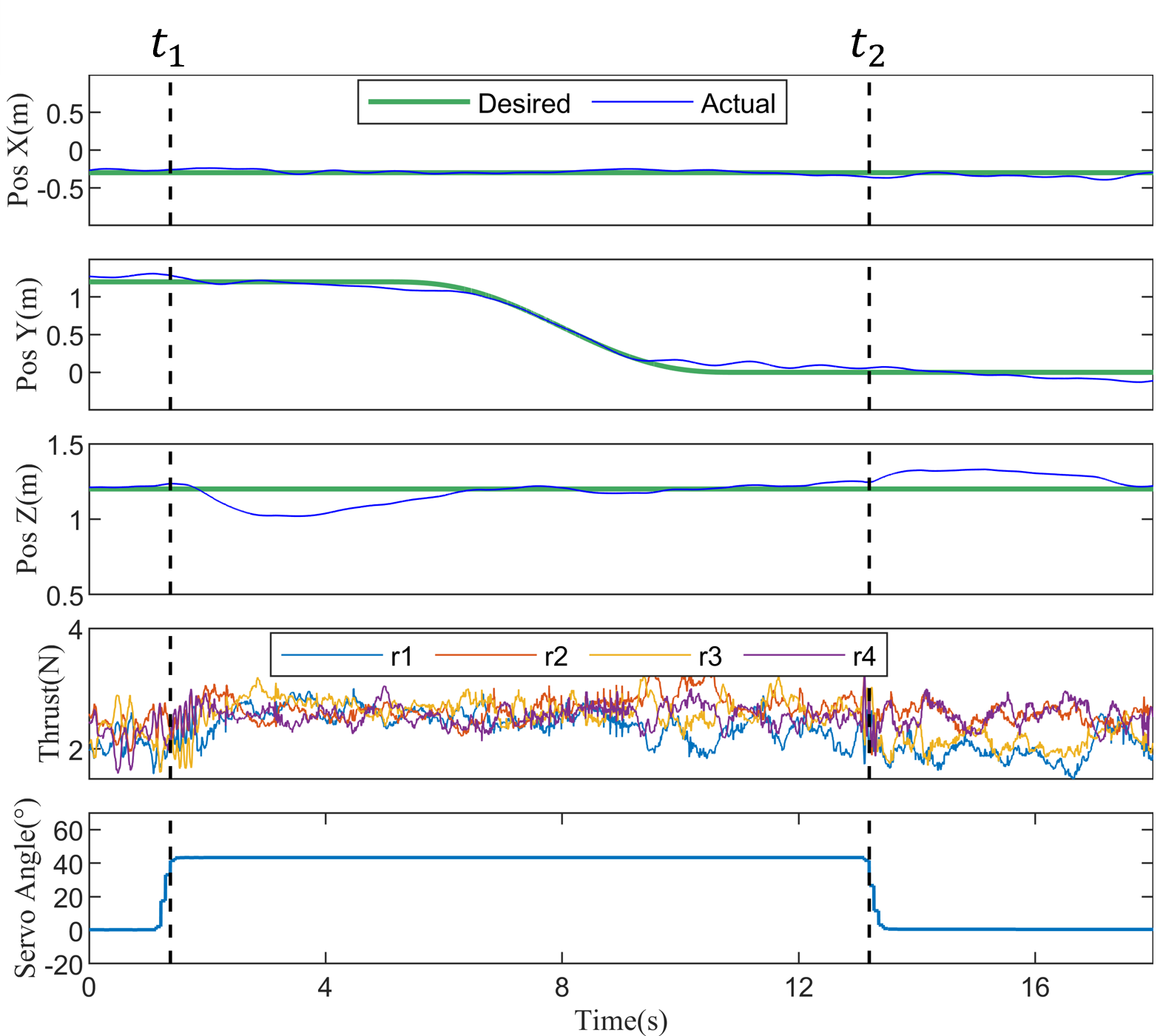}
    \caption{
        Results of the transporting experiment. The quadrotor folded its arms for grasping at $t_1$, maneuvered along $y$ axis to reach the desired position, then stretched the body to release the box at $t_2$. 
        }
    \label{fig:data_grasp}
\end{figure}

\begin{figure*}[t]
    \centering
    \includegraphics[width=\linewidth]{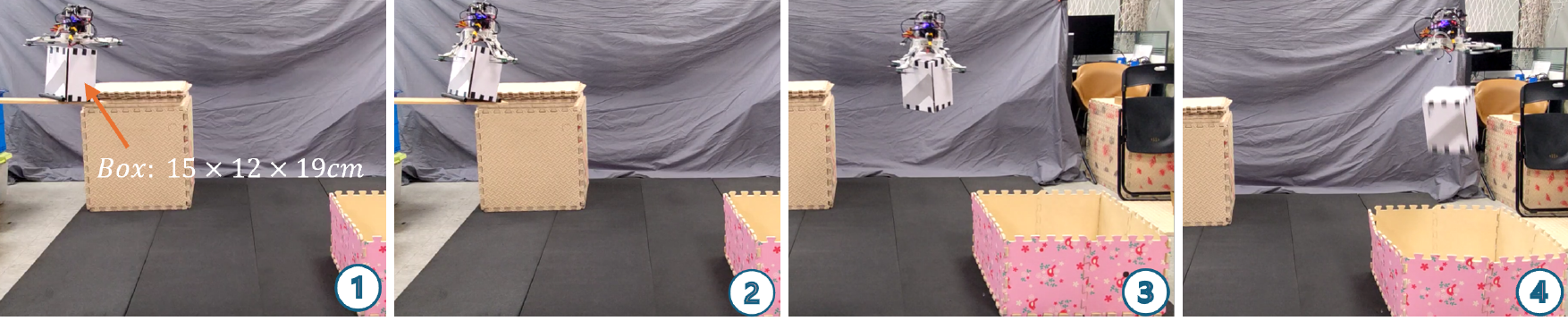}
    \caption{
            The snapshots of grasping and transporting a $77\,\text{g}$ box with the dimensions of $12 \times 15 \times 19\,\text{cm}$. From left to right: \ding{172} hovering, \ding{173} grasping, \ding{174} transporting, \ding{175} releasing.
        }
    \label{fig:video-grasp}
\end{figure*}

Fig. \ref{fig:data_grasp} shows the trajectory during the transporting task. The arms folded at $t_1 = 1.2\,\text{s}$ for grasping and unfolded at $t_2 = 13.2\,\text{s}$ for releasing the object. 
After grasping the object, a noticeable altitude error occurred due to the sudden increase in unknown weight. 
With the adaptive controller automatically adjusting the vehicle's physical properties, the motor thrust increased, quickly eliminating the altitude error.
Similarly, when releasing the object, the sudden reduction in mass caused the quadrotor to drift upward. The controller adjusted motor thrust to restore the desired altitude.
% Benefiting from our adaptive controller, which automatically adjusts the physical properties of the vehicle, the motor thrusts were increased, allowing the altitude error to be quickly eliminated. 
% Similarly, when releasing the object, the sudden reduction in mass caused the quadrotor to drift upward. The controller reduced the motor thrusts to allow the quadrotor to return to the desired altitude. 
Throughout the experiment, tracking remained stable in both the x and y directions, further confirming the proposed controller’s reliability.

\subsection{Traversal of Narrow Spaces}

\begin{figure*}[t]
    \centering
    \includegraphics[width=\linewidth]{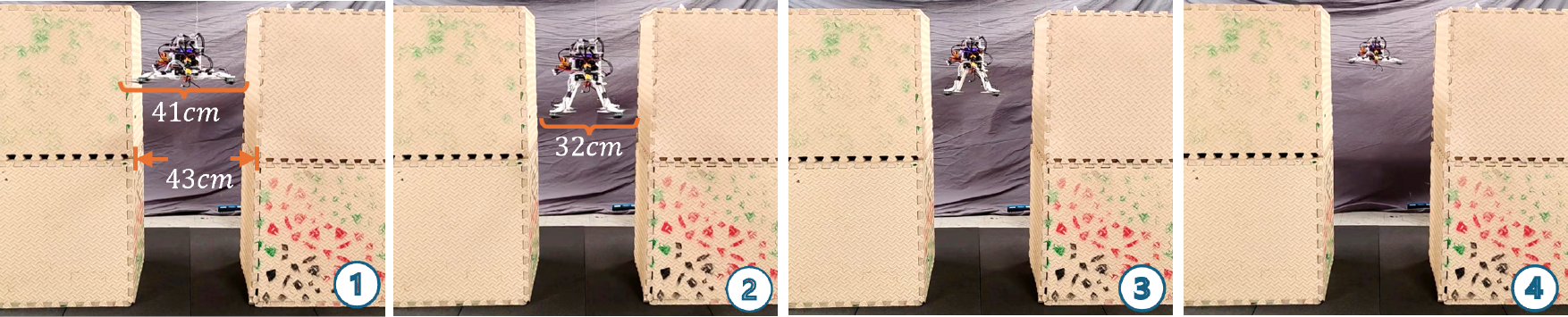}
    \caption{
            The snapshots of passing a narrow gap with the contracted body. From left to right: \ding{172} hovering, \ding{173} contracting, \ding{174} passing, \ding{175} recovering.
        }
    \label{fig:video-passage}
\end{figure*}

Finally, we demonstrate the adaptability of our morphing quadrotor by altering its body configuration to navigate through an obstacle gap.
As shown in Fig. \ref{fig:video-passage},
with the arms unfolded, the total length including the propeller blades is $41\,\text{cm}$, making it challenging to maneuver through the $43\,\text{cm}$-wide passage. 
However, by contracting to the minimal length of $32\,\text{cm}$, the quadrotor can safely traverse the passage with additional tolerance for tracking error.  
This morphing capability enhances the quadrotor’s flexibility in mission execution.

\subsection{Discussions}
% Compared to \cite{bucki2022hinges}, the proposed quadrotor design can actively fold all four arms. It also reduces the frame size and weight compared to our previous work \cite{Xu2024biomimetic}. However, some limitations were observed during the experiments. 
% When the arms are folded, the propellers face downward at the bottom of the quadrotor, leading to a stronger ground effect compared to other types of morphing quadrotors. As a result, the quadrotor exhibited oscillations and even became unstable when approaching the ground with folded arms. Additionally, in the folded state, the close proximity of the rotors and the higher position of the CoG relative to the rotor plane make control more difficult compared to our earlier design in \cite{Xu2024biomimetic}. These factors make it challenging for the proposed quadrotor to perform agile manipulation tasks, such as the dynamic grasping demonstrated in \cite{Xu2024biomimetic}. In future research, we plan to explore the use of multiple morphing quadrotors to address these challenges.

Compared with \cite{jia2023modular,bucki2022hinges}, the proposed quadrotor can actively fold all four arms, and reduce frame size and weight relative to our previous work \cite{Xu2024biomimetic}.
However, some limitations were revealed during the experiments.
When the arms are folded, the propellers face downward at the bottom of the quadrotor, amplifying ground effects and causing oscillations or instability near the ground. 
In addition, the proximity of the rotors and the higher CoG in the folded state make control more challenging than in \cite{Xu2024biomimetic}, limiting agile tasks such as dynamic grasping.
In future research, we plan to explore the use of multiple morphing quadrotors to address these challenges.

\section{Conclusion}

In this paper, we propose a quadrotor capable of morphing by folding its arms vertically. Thanks to the parallelogram structure of the arms and frame, the orientation of the rotors remains unchanged when the arms fold. 
To address the reduced distance between the propellers when the body is contracted, we introduce a disturbance observer into the adaptive controller. 
Comparative experiments confirm that the observer improves control performance. 
Real-world experiments demonstrate that the morphing quadrotor can adapt its size to different spaces and effectively grasp objects. In future work, we will explore the use of multiple morphing quadrotors to achieve more agile manipulation tasks.

% \newpage
% \printbibliography
\bibliography{IEEEtranBST/IEEEabrv,IEEEtranBST/conf_used}

\end{document}